%% file: main.tex
\documentclass[sigconf]{acmart}

\input{include/pkgs}

\input{include/cmds}

\input{include/shorthand}

\AtBeginDocument{%
  }

\setcopyright{none}
\acmYear{}
\acmDOI{}
\acmConference[]{}{}{}
\acmISBN{}
\acmPrice{}

\begin{document}

\title{FASQ: Flexible Accelerated Subspace Quantization for Calibration-Free LLM Compression}

\author{Ye Qiao}
\email{yeq6@uci.edu}
\affiliation{%
  \institution{University of California, Irvine}
  \city{Irvine}
  \state{California}
  \country{USA}
}

\author{Yian Wang}
\email{yianw11@uci.edu}
\affiliation{%
  \institution{University of California, Irvine}
  \city{Irvine}
  \state{California}
  \country{USA}
}

\author{Zhiheng Chen}
\email{zhihenc5@uci.edu}
\affiliation{%
  \institution{University of California, Irvine}
  \city{Irvine}
  \state{California}
  \country{USA}
}

\author{Hyoukjun Kwon}
\email{hyoukjun.kwon.p@gmail.com}
\affiliation{%
  \institution{University of California, Irvine}
  \city{Irvine}
  \state{California}
  \country{USA}
}

\author{Sitao Huang}
\email{sitaoh@uci.edu}
\affiliation{%
  \institution{University of California, Irvine}
  \city{Irvine}
  \state{California}
  \country{USA}
}


\renewcommand{\shortauthors}{Qiao et al.}



\input{sections/00-Abstract}
\maketitle
\input{sections/01-Introduction}
\input{sections/02-Background}
\input{sections/03-FASQ}
\input{sections/04-FASQ_Aceeleration}
\input{sections/05-Experiments}

\input{sections/06-Conclusion}


\input{main.bbl}
\appendix

\end{document}

%% file: include/pkgs.tex
\usepackage{amsmath,amssymb,amsfonts}
\usepackage{algorithmic}
\usepackage{graphicx}
\usepackage[dvipsnames]{xcolor}
\usepackage[all]{nowidow}

\usepackage{array}
\usepackage{multirow}
\usepackage{booktabs}
\usepackage{xspace}
\usepackage[normalem]{ulem}
\usepackage{algorithm}
\usepackage{pifont}
\usepackage{threeparttable}
\usepackage{siunitx}
\usepackage[table]{xcolor}
\usepackage{subcaption}

%% file: include/cmds.tex
\newcommand{\algorithmicparfor}{\textbf{parallel for}}
\newcommand{\PARFOR}[1]{\STATE \algorithmicparfor\ #1\ \textbf{do} \begin{ALC@g}}
\newcommand{\ENDPARFOR}{\end{ALC@g}}
\newcommand{\algorithmiclet}{\textbf{let}}
\newcommand{\LET}[1]{\STATE \algorithmiclet\ #1}

\makeatletter
\renewcommand{\PARFOR}[1]{\STATE \algorithmicparfor\ #1\ \textbf{do} \begingroup \let\COMMENT\ALC@com \begin{ALC@g}}
\renewcommand{\ENDPARFOR}{\end{ALC@g}\endgroup\STATE \textbf{end parallel for}}
\makeatother

\newcommand{\insertFigure}[2]{
    \begin{figure}[htbp]
        \centering
        \includegraphics[width=\linewidth]{figs/#1.pdf}
        \vspace{-7mm}
        \caption{\small #2}
        \vspace{-3mm}
        \label{fig:#1}
    \end{figure}
}

\newcommand{\insertWideFigure}[2]{
    \begin{figure*}[htbp]
        \centering
        \includegraphics[width=\textwidth]{figs/#1.pdf}
        \vspace{-7mm}
        \caption{\small #2}
        \vspace{-4mm}
        \label{fig:#1}
    \end{figure*}
}

%% file: include/shorthand.tex
\newcommand{\name}{\textsc{FASQ}\xspace}

%% file: sections/00-Abstract.tex
\begin{abstract}
Compressing large language models (LLMs) for deployment on commodity GPUs remains challenging: conventional scalar quantization is limited to fixed bit-widths (e.g., 8/4/3-bit), offers only a few discrete compression points, and typically requires calibration data.
We present \textsc{FASQ} (Flexible Accelerated Subspace Quantization), a calibration-free framework that applies product quantization to LLM weight matrices.
By tuning two parameters, sub-vector size and codebook cardinality, \textsc{FASQ} exposes a continuous design space spanning 27--49\% of the original FP16 model size, filling compression gaps that fixed-bit schemes cannot reach.
On Meta-Llama-3-8B, \textsc{FASQ} surpasses 4-bit GPTQ and AWQ in accuracy (67.1--67.7 avg.) at 37--42\% model size, with consistent results on Qwen3-8B and Qwen3.5-9B-Base.
To make product quantization practical at inference time, we design custom CUDA kernels: a LUT-free direct-compute GEMV for decode and an output-stationary double-buffered LUT GEMM for prefill, both with split-K parallelism.
On an RTX~3090, \textsc{FASQ} achieves 45.2\,tok/s decode at effective 4-bit ($2.56\times$ memory reduction) and 51.8\,tok/s at effective 3-bit ($2.80\times$), both surpassing FP16 tensor-core performance (43.9\,tok/s) and delivering $1.6$--$1.8\times$ the throughput of AWQ, $2.2$--$2.5\times$ of GPTQ, and $4.3$--$5.0\times$ of RTN.
\textsc{FASQ} is the only compressed method that accelerates decode beyond FP16, offering calibration-free compression, continuous size--quality trade-offs, and real-time inference on a single consumer GPU.
\end{abstract}

%% file: sections/01-Introduction.tex
\section{Introduction}
Large language models (LLMs) have achieved remarkable performance across a wide range of tasks, yet their deployment is constrained by massive memory footprints and high inference costs.
A single 7--9\,B parameter model requires 14--18\,GB of GPU memory in FP16, often exceeding the capacity of commodity GPUs and limiting real-time serving scenarios.
Post-training quantization (PTQ) has become the predominant solution: compared to other compression methods that require heavy fine-tuning or retraining, PTQ compresses pretrained weights to lower-precision representations without retraining~\cite{gptq,awq,smoothquant,apexq}.

However, existing PTQ methods share three fundamental limitations.
First, scalar quantization is inherently restricted to discrete bit-widths (typically 8, 4, or 3 bits), each yielding a single fixed compression point.
Practitioners seeking intermediate trade-offs (e.g., between the 57\% model size of INT8 and the 36\% of INT4) have no available operating point.
Second, state-of-the-art methods such as GPTQ~\cite{gptq}, AWQ~\cite{awq}, and SmoothQuant~\cite{smoothquant} require representative calibration data to compute Hessian-based corrections, activation-aware scaling factors, or channel migration statistics, respectively.
Obtaining such data may be impractical for proprietary or domain-specific models, and calibration procedures introduce a non-trivial step in the deployment pipeline.
Third, scalar quantization requires explicit dequantization at inference time, reconstructing FP16 weights before each matrix multiplication.
At standard 4-bit precision, custom kernels (e.g., AWQ-W4, GPTQ-W4) amortize this cost effectively, but at non-standard bit-widths where no optimized kernel exists, dequantization overhead can degrade latency by an order of magnitude~\cite{gptq,awq}.

Product quantization (PQ), a classical technique from information retrieval that partitions vectors into sub-vectors and encodes each with a shared codebook, addresses all three limitations.
By varying the sub-vector size and codebook cardinality, PQ exposes a continuous design space of compression rates.
Codebook construction requires only k-means clustering on the weight distribution, eliminating external calibration data.
Most importantly, PQ inference operates directly on compressed representations via table lookup, avoiding dequantization entirely and making latency independent of the compression level.

In this paper, we present \name (\uline{\textbf{F}}lexible \uline{\textbf{A}}ccelerated \uline{\textbf{S}}ubspace \uline{\textbf{Q}}uantization), a post-training compression framework that applies multi-codebook product quantization to the linear layers of LLMs.
\name replaces each FP16 weight matrix with a compact codebook of cluster centroids and a table of uint8 indices, then performs inference directly on these compressed representations: decode uses direct centroid-dot-product accumulation in registers, while prefill uses table-lookup-and-accumulate with shared-memory codebook caching.
Through a systematic design space exploration, we show that \name covers compression rates from 27\% to 49\% of FP16 model size with fine granularity, filling the gap between fixed-bit scalar methods.

On Meta-Llama-3-8B, \name matches 8-bit quality (68.2 average accuracy across five zero-shot tasks) at 49\% model size and exceeds 4-bit AWQ/GPTQ (67.1--68.2 avg.) at 37--42\% model size, all without calibration data. These results generalize consistently to Qwen3-8B and Qwen3.5-9B-Base.
To make PQ inference practical, we develop custom CUDA kernels: a direct-compute GEMV kernel for decode that eliminates shared-memory lookup tables entirely, and a split-K GEMM kernel for prefill that maximizes GPU occupancy.
The resulting engine achieves 45.2\,tok/s decode throughput on an RTX 3090, surpassing FP16 (43.9\,tok/s) at $2.56\times$ less GPU memory.
Because \name operates directly on compressed representations rather than dequantizing to FP16, its decode latency actually \emph{improves} at higher compression, unlike scalar methods where non-standard bit-widths require entirely different kernel implementations.

Our contributions are as follows:
\begin{itemize}
    \item We introduce \name, a calibration-free post-training compression framework that applies product quantization to LLM weights, exposing a continuous design space of compression vs.\ quality trade-offs unreachable by fixed-bit scalar methods.
    \item We evaluate \name across three modern LLM families (Llama-3, Qwen3, Qwen3.5) and show that it achieves accuracy competitive with state-of-the-art INT4 and INT8 quantization without any calibration data.
    \item We design two specialised CUDA kernels (a LUT-free direct-compute GEMV for decode and an output-stationary double-buffered LUT GEMM for prefill), making \name the only compressed method whose decode throughput surpasses FP16 tensor-core performance while reducing memory by $2.56$--$2.80\times$.
    \item We show that \name's decode latency \emph{improves} at higher compression (45.2\,tok/s at eff.\ 4-bit, 51.8\,tok/s at eff.\ 3-bit), inverting the latency penalty of scalar methods and delivering $1.6$--$1.8\times$ the throughput of AWQ, $2.2$--$2.5\times$ of GPTQ, and $4.3$--$5.0\times$ of RTN.
\end{itemize}

%% file: sections/02-Background.tex
\section{Background and Motivation}

\subsection{Scalar Quantization for LLMs}
Scalar quantization reduces the precision of individual weights from floating-point to a lower bit-width integer representation.
Given a weight tensor $w$, the uniform affine mapping to $b$-bit integers is:
\begin{equation}
    Q(w) = \text{clamp}\!\left(\left\lfloor \frac{w}{s} \right\rceil + z,\; 0,\; 2^b{-}1\right)
    \label{eq:sq}
\end{equation}
where $s = (\max(w) - \min(w))\,/\,(2^b{-}1)$ is the scale factor and $z = \lfloor -\min(w)/s \rceil$ is the zero-point.
While simple and hardware-friendly, this mapping suffers large quantization errors when outlier values dominate the dynamic range, which is common in LLM attention and feed-forward layers.

To mitigate outlier-induced error, state-of-the-art post-training quantization (PTQ) methods introduce calibration-based preprocessing.
GPTQ~\cite{gptq} performs layer-wise quantization guided by a Hessian approximation computed from a calibration set, greedily rounding weights to minimize reconstruction error.
AWQ~\cite{awq} identifies salient weight columns via activation statistics and rescales them to fit the INT4 range more tightly.
SmoothQuant~\cite{smoothquant} migrates quantization difficulty from activations to weights through per-channel scaling, enabling joint W8A8 quantization.
These methods achieve strong quality at INT4 and INT8 but share three structural limitations.
First, each bit-width yields a single fixed compression point; no intermediate operating points exist between, e.g., the 57\% model size of INT8 and the 36\% of INT4.
Second, all three require representative calibration data (Hessian statistics, activation distributions, or channel scaling factors) which may be unavailable for proprietary or domain-specific models.
Third, scalar quantization requires explicit dequantization at inference time: integer weights must be reconstructed to FP16 before each matrix multiplication, and this cost is amortized only when a dedicated kernel exists for that bit-width.
Group quantization partially addresses the first limitation by applying per-group scale factors within a fixed bit-width, but the compression range remains narrow and calibration is still required.

\subsection{Product Quantization}\label{sec:bg-pq}
Vector quantization (VQ)~\cite{vq} quantizes entire vectors rather than individual scalars.
A codebook of $K$ centroids is learned via k-means, and each input vector is replaced by the index of its nearest centroid.
By capturing correlations within vectors that element-wise methods ignore, VQ yields lower reconstruction error at equivalent storage.

Product quantization (PQ)~\cite{pq} scales VQ to high-dimensional spaces by partitioning each vector into $N_{ss}$ sub-vectors and maintaining an independent codebook per partition.
The composite codebook has $K^{N_{ss}}$ implicit entries while storing only $K \times N_{ss}$ centroids, making PQ exponentially more expressive than a single-codebook VQ of the same size.
Codebook learning requires only k-means on the data itself, with no external labels or calibration inputs, and the resulting representation is a compact pair of tables: centroids (FP16) and indices (uint8).

Several works have applied PQ or lookup-based computation to neural network inference.
Maddness~\cite{maddness} replaces multiply-accumulate operations with learned hash lookups, and PECAN~\cite{pecan} uses content-addressable memory for mapping, but both target activations rather than weight compression.
DPQ~\cite{dpq}, LUT-NN~\cite{lut-nn}, and PQA~\cite{pqa} apply PQ to network weights or LLM inference but rely on na\"ive dequantization without custom GPU kernels, incurring substantial inference overhead.
LUT-DLA~\cite{lut-dla} adds a design-space search for hardware co-optimization but targets FPGA rather than commodity GPUs.


Orthogonally, table-lookup engines on edge FPGAs~\cite{tellme,pdswap,cobra} and efficient NAS for constrained platforms~\cite{micronas,monas,tgnas,bnn_rimc,spatial_arch} target custom hardware or extreme quantization, while quantization-induced positional encoding degradation~\cite{qroar,rope_rescaling} further motivates calibration-free approaches.
None of these works provide flexible, GPU-efficient compression with a continuous quality and size tradeoff.

%% file: sections/03-FASQ.tex
\section{\name}\label{sec:fasq}
This section describes the \name quantization procedure (\autoref{sec:fasq-quant}), reconstruction-free inference (\autoref{sec:fasq-inference}), and the resulting design space (\autoref{sec:dse}).

\subsection{\name Methodology}\label{sec:fasq-method}
\subsubsection{Quantization}\label{sec:fasq-quant}
\input{algorithms/apq-train}
\insertFigure{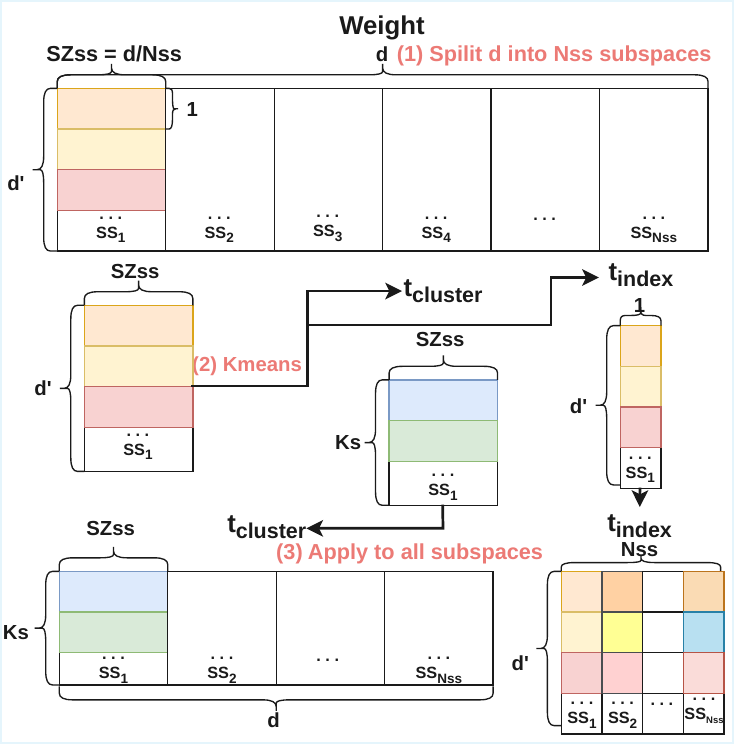}{Example \name quantization process with $N_{ss}=4$. \ding{172}: The weight matrix is split into $N_{ss}$ subspaces along the chosen dimension; \ding{173}: K-means clustering is applied independently within each subspace; \ding{174}: Centroids and indices are stored as the compressed representation.}

\name partitions each weight matrix $W \in \mathbb{R}^{d \times d'}$ into $N_{ss}$ subspaces along a chosen \emph{subspace dimension} $dim_{ss}$.
The partitioning can be applied along either the row ($d$) or column ($d'$) axis; we denote this choice by a binary variable $dim$ ($0$ for rows, $1$ for columns).
The complementary axis is the \emph{datapoint dimension} $dim_{dp}$.
Each subspace slice has shape:
\begin{equation}
    W_{ss} \in 
    \begin{cases} 
    \mathbb{R}^{\frac{d}{N_{ss}} \times d'}, & \text{if } dim = 0 \\[4pt]
    \mathbb{R}^{d \times \frac{d'}{N_{ss}}}, & \text{if } dim = 1
    \end{cases}
    \label{eq:subspace-dimension}
\end{equation}
with subspace size $SZ_{ss} = dim_{ss} / N_{ss}$.

Within each subspace, we run k-means with $K_s$ clusters along $dim_{dp}$, producing a centroid table $T_{\text{cluster}}^{ss} \in \mathbb{R}^{K_s \times SZ_{ss}}$ and an index vector $T_{\text{index}}^{ss} \in \{0,\ldots,K_s{-}1\}^{dim_{dp}}$.
Aggregating over all subspaces yields the full codebook $T_{\text{cluster}} \in \mathbb{R}^{N_{ss} \times K_s \times SZ_{ss}}$ and the index table $T_{\text{index}} \in \{0,\ldots,K_s{-}1\}^{N_{ss} \times dim_{dp}}$.
Because k-means operates solely on the weight distribution, \textbf{no calibration data is required}.
The complete procedure is given in \autoref{alg:fasq-train} and illustrated in \autoref{fig:apq_iccad26}.

\subsubsection{Inference}\label{sec:fasq-inference}

A na\"ive approach reconstructs the full weight matrix via index lookups and then calls a standard matrix multiplication.
However, reconstruction temporarily materializes the entire FP16 weight matrix, negating the memory savings of quantization.

We instead perform \textbf{reconstruction-free} inference that operates directly on the codebook and index tables without ever materializing the weight matrix.
The core observation is that the output of a PQ-compressed linear layer can be decomposed as:
\begin{equation}
    o_j = \sum_{ss=1}^{N_{ss}} \text{dot}\!\bigl(\,x_{ss},\; T_{\text{cluster}}[ss,\, T_{\text{index}}[ss, j]]\,\bigr)
    \label{eq:lut-accumulate}
\end{equation}
where $x_{ss}$ is the input slice corresponding to subspace $ss$.
Each term requires only a single centroid fetch (indexed by $T_{\text{index}}$) and a small dot product of dimension $SZ_{ss}$; no full row of the weight matrix is ever formed.

We specialize the implementation into two kernels tailored to the two LLM inference phases:
\begin{itemize}
    \item \textbf{Decode (GEMV):} During autoregressive token generation, the input is a single vector ($B{=}1$).
    Each thread owns one output element and directly computes \autoref{eq:lut-accumulate} by iterating over subspaces, fetching the relevant centroid on the fly, and accumulating the dot product in a register.
    No shared-memory lookup table is needed, because each thread uses only one of the $K_s$ centroids per subspace.
    \item \textbf{Prefill (GEMM):} During prompt processing, many input tokens ($L \gg 1$) share the same weight.
    Here, precomputing a shared-memory lookup table $\text{LUT}_{ss}[k] = x_l \cdot T_{\text{cluster}}[ss, k]^\top$ for all $K_s$ centroids amortizes the cost across the $L$ output rows, and each thread accumulates results for multiple output features via indexed gathers from the LUT.
\end{itemize}
Both approaches are inherently dequantization-free: the compressed codebook and index tables are consumed directly, with no intermediate FP16 weight materialization.
The detailed kernel designs and CUDA-level optimisations are presented in \autoref{sec:acceleration}.

\subsection{\name Design Space}\label{sec:dse}
\name targets the linear layers of transformer models, which account for the vast majority of parameters.
We assume the baseline model stores weights in FP16 (16 bits per element).

\name compresses each layer by replacing the full $dim_{ss} \times dim_{dp}$ weight matrix with two compact structures:
(1)~a \emph{codebook} of $N_{ss} \times K_s$ centroids, each of dimension $SZ_{ss}$, shared across all $dim_{dp}$ datapoints; and
(2)~an \emph{index table} that stores one $\lceil \log_2 K_s \rceil$-bit integer per datapoint per subspace instead of $SZ_{ss}$ FP16 values.
The per-layer footprint is therefore:
\begin{align}
    S_{\text{fasq}}^{\text{layer}}
      &= \underbrace{16 \times K_s \times dim_{ss}}_{\text{codebook (shared)}}
       + \underbrace{\lceil \log_2 K_s \rceil \times N_{ss} \times dim_{dp}}_{\text{index table}}
    \label{eq:reduced-model-size}
\end{align}
The codebook cost is independent of $dim_{dp}$ and therefore negligible at typical model dimensions (e.g.\ $<$1\% of layer size at $K_s{=}256$, $d{=}4096$).
Compression is dominated by the index table, which replaces each FP16 weight group with a single $\lceil \log_2 K_s \rceil$-bit index.
For a given model and choice of $dim$, the layer size is determined entirely by $K_s$ and $N_{ss}$.

The model-level size ratio aggregates all \name-quantized linear layers together with uncompressed auxiliary components (embeddings, LayerNorm):
\begin{equation}
    \text{Size\%} = \frac{S_{\text{fasq}}}{S_{\text{base}}} \times 100\%
    \label{eq:compression-rate}
\end{equation}
We additionally apply deduplication by removing repeated centroids within each subspace; hence \autoref{eq:reduced-model-size} is an upper bound.
In our tables, the effective bitwidth \textbf{\#W} reports only the index rate $\lceil \log_2 K_s \rceil / SZ_{ss}$, excluding codebook overhead, analogous to how integer quantization methods report precision without counting group-wise scaling factors.

We define the \name design space as:
\begin{equation}
    \text{Design Space} = \{K_s,\; N_{ss},\; dim\}
    \label{eq:design-space}
\end{equation}
operating on FP16 weights, the standard deployment format for pretrained LLMs.
By sweeping $SZ_{ss}$ from 1 (element-wise k-means) to $dim_{ss}$ (full vector quantization) and $K_s$ from 1 to $dim_{dp}$, \name traces a continuous Pareto front of compression rate versus model quality (\autoref{sec:dse-results}).

%% file: algorithms/apq-train.tex
\begin{algorithm}[ht]
    \caption{\name Quantization Process}
    \label{alg:fasq-train}
    \begin{algorithmic}[1]
        \REQUIRE Weight matrix $W$ with shape $(d, d')$, number of clusters $K_s$, number of subspaces $N_{ss}$, subspace dimension $dim$.
        \ENSURE Cluster table $T_{\text{cluster}}$, index table $T_{\text{index}}$
        \LET{$dim_{dp}$ be the data point dimension.}
        \STATE $dim_{ss} \gets d \text{ if } dim \gets 0 \text{ else } d'$, $dim_{dp}$ be the number of data points
        \STATE $SZ_{ss} \gets \frac{dim_{ss}}{N_{ss}}$
        \STATE Initialize $T_{\text{cluster}}\in\mathbb{R}^{N_{ss}, K_s, SZ_{ss}}$, $T_{\text{index}}\in\mathbb{R}^{N_{ss}, n_{dp}}$
        \PARFOR{$ss_i \gets 1$ to $N_{ss}$}
            \STATE $(C, I) \gets \text{KMeans}(W_{ss_i}, K_s)$
            \STATE $T_{\text{cluster}}[ss_i] \gets C$
            \STATE $T_{\text{index}}[ss_i] \gets I$
        \ENDPARFOR
        \RETURN $T_{\text{cluster}}$, $T_{\text{index}}$
    \end{algorithmic}
\end{algorithm}

%% file: sections/04-FASQ_Aceeleration.tex
\section{\name Acceleration}\label{sec:acceleration}
LLM inference consists of two phases: \emph{prefill}, which processes the prompt via a batched GEMM, and \emph{decode}, which generates tokens one at a time via GEMV.
The two phases impose different parallelism requirements, so we design a dedicated CUDA kernel for each.
We present the designs of both kernels, an ablation study of the key optimisations, and a sensitivity analysis of \name parameters.
All microbenchmark measurements are on a single $4096{\times}4096$ linear layer, RTX~3090.

A natural approach to PQ inference is to precompute a \emph{lookup table} (LUT) of all $K_s$ dot products per subspace in shared memory, then gather results by index (\autoref{eq:lut-accumulate}).
Whether this pays off depends on the batch size: at $B{=}1$ each thread uses only one of the $K_s$ entries, wasting the table; at large $L$, many tokens amortize the build cost.
This tradeoff drives our two-kernel design.

\subsection{GEMV Kernel for Decode}\label{sec:gemv-kernel}
\input{algorithms/apq-gemv}
\insertWideFigure{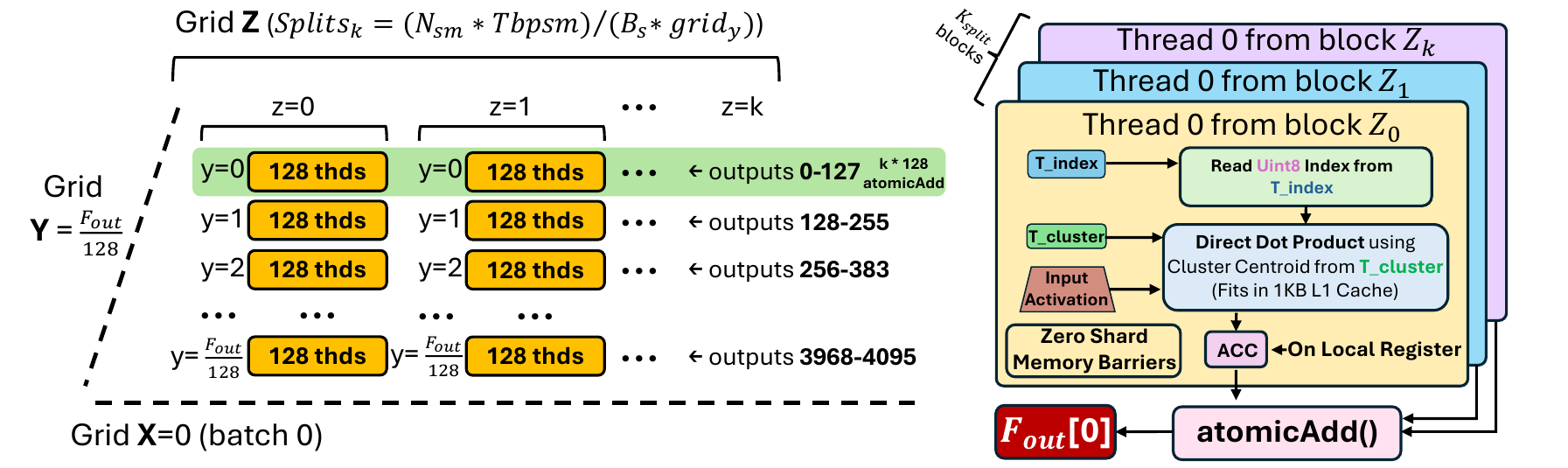}{\name GEMV kernel design for autoregressive decode ($B{=}1$). Left: the 3D grid layout, where grid-$y = \lceil F_{\text{out}} / 128 \rceil$ partitions outputs into 128-thread blocks and grid-$z$ implements Split-K by dividing $N_{ss}$ subspaces across $K_{\text{split}}$ blocks (auto-tuned for ${\sim}8$ blocks/SM). Right: per-thread computation, each thread reads a uint8 index from $T_{\text{index}}$, performs a direct dot product with the corresponding centroid from $T_{\text{cluster}}$ (fitting in 1\,KB L1 cache), and accumulates in a local register with zero shared memory or barriers. Partial sums from all $K_{\text{split}}$ blocks merge via a single \texttt{atomicAdd} per output element.}

During autoregressive decode, the input is a single token.
A na\"ive LUT approach builds $K_s$ partial dot products per subspace in shared memory, but each thread consumes only 1 of the $K_s$ entries; at $K_s{=}256$, \textbf{99.6\% of the LUT is wasted}.

Our GEMV kernel (\autoref{alg:fasq-gemv}) eliminates the LUT entirely, as illustrated in \autoref{fig:Fig_gemv_cropped}.
Each thread owns a single output element $o_j$ and iterates over its assigned subspaces.
For each subspace $ss$, the thread reads the index $k = T_{\text{index}}[ss, j]$ (1~byte), fetches the corresponding centroid $T_{\text{cluster}}[ss, k]$ ($SZ_{ss} \times 2$~bytes), computes the dot product with the input slice using $SZ_{ss}$ fused multiply-add (FMA) operations, and accumulates the result in a register (\autoref{fig:Fig_gemv_cropped}, right).
After all subspaces are processed, a single \texttt{atomicAdd} writes the partial sum to global memory.

\subsubsection*{Split-K parallelism}
To maximise GPU occupancy, we partition the $N_{ss}$ subspaces across the grid's z-dimension (\autoref{fig:Fig_gemv_cropped}, left).
The auto-tuner targets ${\sim}8$ blocks per SM, yielding $K_{\text{split}} = \lceil (N_{SM} \times 8) / (B_s \times \lceil F_{\text{out}} / 128 \rceil) \rceil$ splits (${\sim}21$ on RTX~3090).
Each thread performs a single \texttt{atomicAdd} at the end of its subspace chunk, compared with $N_{ss}$ atomics per element in a subspace-stationary design.

\subsubsection*{Key properties}
This design uses \textbf{zero shared memory} and \textbf{zero synchronisation barriers}, enabling up to 12 resident blocks per SM (versus 1 for the 32\,KB LUT design).
At $SZ_{ss}{=}2$, the centroid is a single \texttt{half2} load; the entire subspace codebook ($K_s {\times} SZ_{ss} {\times} 2 = 1$\,KB) fits in L1 and is reused across the warp.
The dominant cost is reading the index table: $N_{ss} {\times} F_{\text{out}} {\times} 1$\,B $= 8$\,MB vs.\ cuBLAS FP16's $32$\,MB weight read, yielding a $4\times$ memory traffic reduction that explains why our kernel ($32\,\mu$s) outperforms cuBLAS ($45\,\mu$s).

\subsection{GEMM Kernel for Prefill}\label{sec:gemm-kernel}
\input{algorithms/apq-gemm}

During prefill, $L$ input tokens share the same weight.
Building a $K_s$-entry LUT per subspace now pays off: the $L$ tokens each reuse the same table, amortizing the build cost by a factor of $L$.
Our GEMM kernel assigns one sequence position to each block row (grid-$x$) and iterates serially over subspaces.
For each subspace, a $K_s$-entry LUT is built cooperatively by all threads (each thread computes $\lceil K_s / 256 \rceil$ entries), then each thread looks up 16 output features from the LUT and accumulates results in registers.

\subsubsection*{Split-K along $N_{ss}$}
At short sequences ($L{=}32$--128), we split the $N_{ss}$-loop across the grid z-dimension to increase block count.
At $L{=}128$ the auto-tuner selects ${\sim}4$ splits, boosting blocks from 128 to 512 ($6.2$/SM) and reducing latency by $1.35\times$; at large $L$, splits are disabled with zero overhead.
\subsubsection*{Double-buffered LUT}
Two LUT slots reside in shared memory: while accumulators read from \texttt{lut\_cur}, \texttt{lut\_nxt} is built for the next subspace, requiring only one \texttt{\_\_syncthreads()} per iteration.
Each slot is padded to $K_s{+}1{=}257$ floats (2\,KB for the pair) for bank-conflict avoidance.

\subsection{Kernel Design Ablation}\label{sec:opt-progression}

\autoref{tab:kernel-versions} ablates each optimisation cumulatively on a single $4096{\times}4096$ layer.

\begin{table}[t]
    \centering
    \caption{Kernel design ablation on RTX~3090 ($4096{\times}4096$ layer). Each row adds one optimisation cumulatively.}
    \label{tab:kernel-versions}
    \setlength{\tabcolsep}{3pt}
    \renewcommand{\arraystretch}{1.1}
    \resizebox{\columnwidth}{!}{%
    \begin{tabular}{l|l|r|r}
        \toprule
        & \textbf{Optimisation} & \textbf{GEMV} & \textbf{GEMM} \\
                       &                     & \textbf{($B{=}1$, $\mu$s)} & \textbf{($L{=}128$, ms)} \\
        \midrule
        (a) & cuBLAS FP16 (tensor core)        & 45  & 0.08 \\
        (b) & Subspace-stationary LUT (both)   & 142 & 5.11 \\
        (c) & + Output-stationary; half2, fused f2h & 94  & 2.75 \\
        (d) & + LUT-free direct compute, split-K (GEMV) & \textbf{32}  & -- \\
        (e) & + Split-K, double-buffered LUT (GEMM) & --  & \textbf{2.02} \\
        \bottomrule
    \end{tabular}}
\end{table}

The baseline PQ kernel (b) uses a shared-memory LUT with subspace-stationary parallelism, where each block owns one subspace and iterates over all output elements.
Switching GEMM to an output-stationary design (each thread owns output features, iterates over subspaces) and adding \texttt{half2} vectorised loads, fused float-to-half conversion, and PyTorch's caching allocator (c) yields a $1.5\times$ GEMV and $1.9\times$ GEMM speedup.
Eliminating the LUT entirely and adding split-K parallelism along $N_{ss}$ (d) achieves a further $2.9\times$ GEMV speedup, bringing decode latency \textbf{below} cuBLAS FP16; the GEMM kernel retains its LUT since the cost is amortized across $L$ tokens.
Finally, GEMM split-K (also along $N_{ss}$, but retaining the LUT) and double-buffered LUT (e) provide a $1.35\times$ prefill speedup at short sequence lengths.

\subsection{Sensitivity Analysis}\label{sec:kernel-sensitivity}

\subsubsection*{Codebook size ($K_s$)}
\begin{figure*}[t]
    \centering
    \begin{subfigure}[t]{0.32\textwidth}
        \centering
        \includegraphics[width=\linewidth]{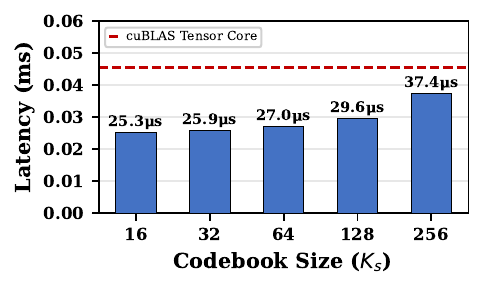}
        \caption{GEMV decode latency vs.\ $K_s$ ($B{=}1$, $SZ_{ss}{=}2$). All $K_s$ values beat cuBLAS.}
        \label{fig:gemv-ks-sensitivity}
    \end{subfigure}\hfill
    \begin{subfigure}[t]{0.32\textwidth}
        \centering
        \includegraphics[width=\linewidth]{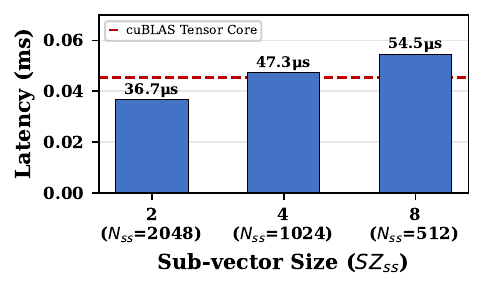}
        \caption{GEMV decode latency vs.\ $SZ_{ss}$ ($B{=}1$, $K_s{=}256$). $SZ_{ss}{=}2$ is optimal.}
        \label{fig:gemv-szss-sensitivity}
    \end{subfigure}\hfill
    \begin{subfigure}[t]{0.32\textwidth}
        \centering
        \includegraphics[width=\linewidth]{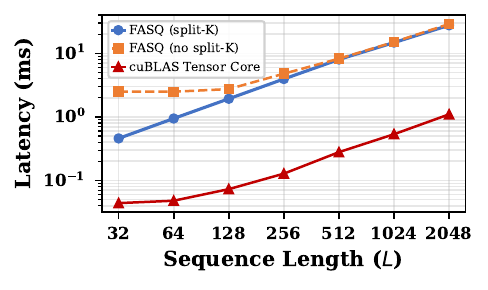}
        \caption{GEMM prefill latency vs.\ $L$ ($SZ_{ss}{=}2$, $K_s{=}256$). Split-K helps most at short $L$.}
        \label{fig:gemm-seq-scaling}
    \end{subfigure}
    \caption{Sensitivity analysis of \name kernels on a single $4096{\times}4096$ layer (RTX~3090).}
    \label{fig:sensitivity}
\end{figure*}

As $K_s$ varies from 16 to 256 (\autoref{fig:gemv-ks-sensitivity}), GEMV latency rises only from 25.3 to 37.4\,$\mu$s ($1.48\times$) because the dominant cost, reading the index table ($N_{ss} {\times} F_{\text{out}}$ bytes), is independent of $K_s$.
\textbf{All $K_s$ values remain faster than cuBLAS FP16} (45.5\,$\mu$s), so compression quality can be tuned freely.
\subsubsection*{Sub-vector size ($SZ_{ss}$)}
$SZ_{ss}{=}1$ doubles the index table to 16\,MB and disables \texttt{half2} vectorisation ($134.7\,\mu$s, $3.0\times$ cuBLAS).
$SZ_{ss}{=}2$ halves index traffic and enables half2, reaching 36.7\,$\mu$s ($0.81\times$ cuBLAS); larger values give diminishing returns. We use $SZ_{ss}{=}2$ throughout.

\subsubsection*{Sequence length scaling}
Split-K provides the largest GEMM benefit at short sequences ($4.5\times$ at $L{=}32$), tapering to $1.01\times$ at $L \geq 512$ where blocks already saturate the GPU.
The \name GEMM is ${\sim}26$--$28\times$ slower than cuBLAS FP16 at large $L$ (tensor-core gap); however, prefill runs once per request while decode dominates interactive serving (\autoref{fig:gemm-seq-scaling}).
\subsubsection*{Batch size}
At $B{=}1$, our GEMV kernel is $1.2\times$ \textbf{faster} than cuBLAS FP16, as the $4\times$ memory traffic reduction outweighs the lack of tensor cores.
As $B$ grows, cuBLAS scales more favourably via tensor-core GEMM; an adaptive dispatch switching kernels at larger $B$ is left for future work.

%% file: algorithms/apq-gemv.tex
\begin{algorithm}[t]
    \caption{\name GEMV Kernel (Decode, $B{=}1$)}
    \label{alg:fasq-gemv}
    \begin{algorithmic}[1]
        \REQUIRE $x \in \mathbb{R}^{F_{\text{in}}}$, $T_{\text{cluster}} \in \mathbb{R}^{N_{ss} \times K_s \times SZ_{ss}}$
        \REQUIRE $T_{\text{index}} \in \{0,\ldots,K_s{-}1\}^{N_{ss} \times F_{\text{out}}}$
        \ENSURE $o \in \mathbb{R}^{F_{\text{out}}}$
        \STATE $o \gets \mathbf{0}$;\ auto-tune \textit{nss\_splits} for ${\sim}8$ blks/SM
        \PARFOR{thread $(j, z)$: $j \in [0, F_{\text{out}})$, $z \in [0, \textit{nss\_splits})$}
            \STATE $\textit{acc} \gets 0$ \hfill $\triangleright$ register
            \FOR{$ss$ in split $z$'s subspace range}
                \STATE $k \gets T_{\text{index}}[ss, j]$ \hfill $\triangleright$ 1\,B index
                \STATE $c \gets T_{\text{cluster}}[ss, k]$ \hfill $\triangleright$ centroid
                \STATE $\textit{acc} \mathrel{+}= \text{dot}(x_{ss},\, c)$ \hfill $\triangleright$ $SZ_{ss}$ FMAs
            \ENDFOR
            \STATE $\texttt{atomicAdd}(o[j],\, \textit{acc})$ \hfill $\triangleright$ 1 atomic/thread
        \ENDPARFOR
    \end{algorithmic}
\end{algorithm}

%% file: algorithms/apq-gemm.tex
\begin{algorithm}[t]
    \caption{\name GEMM Kernel (Prefill, $L > 1$)}
    \label{alg:fasq-gemm}
    \begin{algorithmic}[1]
        \REQUIRE $X \in \mathbb{R}^{L \times F_{\text{in}}}$, $T_{\text{cluster}}$, $T_{\text{index}}$
        \ENSURE $O \in \mathbb{R}^{L \times F_{\text{out}}}$
        \STATE $O \gets \mathbf{0}$;\ auto-tune \textit{nss\_splits} for ${\sim}6$ blks/SM
        \PARFOR{block $(l, z)$: $l \in [0, L)$, $z \in [0, \textit{nss\_splits})$}
            \STATE Alloc double-buffered $\textit{lut}_0, \textit{lut}_1 \in \mathbb{R}^{K_s}$ in shmem
            \STATE Each thread owns $\textit{acc}[0{:}F_{\text{pt}}{-}1] \gets \mathbf{0}$ \hfill $\triangleright$ $F_{\text{pt}}{=}16$
            \STATE \textbf{Build} $\textit{lut}_0[k] \gets \text{dot}(X[l]_{ss_0},\, T_{\text{cluster}}[ss_0, k])$, $\forall k$ \hfill 
            \STATE \texttt{\_\_syncthreads()}
            \FOR{each subspace $ss$ in split $z$'s range}
                \STATE $\textit{cur} \gets \textit{lut}_{i \bmod 2}$;\ $\textit{nxt} \gets \textit{lut}_{(i{+}1) \bmod 2}$
                \STATE \textbf{Build} $\textit{nxt}$ for next subspace \hfill $\triangleright$ overlaps reads
                \FOR{$p = 0$ \textbf{to} $F_{\text{pt}}-1$}
                    \STATE $\textit{acc}[p] \mathrel{+}= \textit{cur}[\, T_{\text{index}}[ss, j_p]\,]$ \hfill $\triangleright$ LUT gather
                \ENDFOR
                \STATE \texttt{\_\_syncthreads()} \hfill $\triangleright$ 1 barrier/subspace
            \ENDFOR
            \STATE Store: direct fp16 if $\textit{nss\_splits}{=}1$, else \texttt{atomicAdd}
        \ENDPARFOR
    \end{algorithmic}
\end{algorithm}

%% file: sections/05-Experiments.tex
\section{Experiments}\label{sec:experiments}
We evaluate \name on an NVIDIA RTX 3090 GPU, measuring model quality, compressed model size, and end-to-end inference latency with our custom CUDA kernels.

\subsection{Experimental Settings}

\subsubsection{Baselines}
We compare against five post-training quantization methods:
GPTQ~\cite{gptq} (Hessian-guided INT4/INT3),
AWQ~\cite{awq} (activation-aware weight quantization),
SmoothQuant~\cite{smoothquant} (joint weight-activation quantization at W8A8, W6A6, and W4A4),
QuIP~\cite{quip} (incoherence processing before quantization),
and RTN (round-to-nearest, no calibration).
All baselines except RTN require calibration data; \name does not.

\subsubsection{LLM Workloads}
We evaluate on three modern model families: Meta-Llama-3-8B~\cite{llama3}, Qwen3-8B~\cite{qwen3}, and Qwen3.5-9B-Base~\cite{qwen3}.
For scalability analysis, we additionally evaluate on LLaMA-2~\cite{llama2} 7B and 13B.
We measure perplexity on the WikiText-2 test set~\cite{wikitext} and zero-shot accuracy on five commonsense reasoning tasks: ARC-easy and ARC-challenge~\cite{arc}, HellaSwag~\cite{hellaswag}, PIQA~\cite{piqa}, and WinoGrande~\cite{winogrande}.

\subsubsection{Configurations}
\name is parameterized by sub-vector size $SZ_{ss}$ and codebook size $K_s$, written as $SZ_{ss}$-$K_s$ (e.g., 2-256 denotes $SZ_{ss}{=}2$, $K_s{=}256$).

\subsubsection{Methodology}
We reproduced baselines using official implementations of GPTQ\footnote{\url{https://github.com/ist-daslab/gptq}}, AWQ\footnote{\url{https://github.com/mit-han-lab/llm-awq}}, and SmoothQuant\footnote{\url{https://github.com/mit-han-lab/smoothquant}}.
All evaluations use lm-evaluation-harness~\cite{lm-eval}.
Compression rates are measured from stored model weight size only, excluding activations and KV caches.
End-to-end latency is measured on Meta-Llama-3-8B with prompt length 128 and generation length 128.
All latency measurements use CUDA event timers with warmup.

\subsection{Design Space Exploration}\label{sec:dse-results}
As discussed in \autoref{sec:dse}, \autoref{eq:reduced-model-size} defines the memory footprint of a single \name-quantized linear layer.
We sweep $SZ_{ss} \in \{1,2,4,8\}$ and $K_s \in \{64,128,256,512,1024,2048\}$ on Meta-Llama-3-8B to map the full design space.
All experiments use $dim{=}0$ (quantization along the output dimension), which we found consistently outperforms $dim{=}1$ across all configurations.

\begin{figure}[t]
    \centering
    \includegraphics[width=\linewidth]{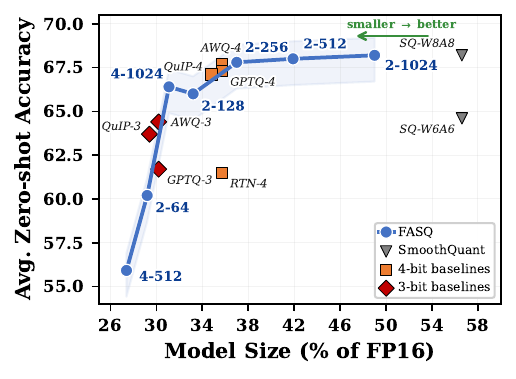}
    \vspace{-6mm}
    \caption{Pareto front of \name design space on Meta-Llama-3-8B. Each point is one ($SZ_{ss}$, $K_s$) configuration. \name with $SZ_{ss}{\leq}2$ traces a smooth front that matches or exceeds scalar baselines across the full size range.}
    \label{fig:dse-pareto}
    \vspace{-3mm}
\end{figure}

\autoref{fig:dse-pareto} plots each configuration by model size (\% of FP16) versus average zero-shot accuracy, with scalar quantization baselines overlaid for reference.
Configurations with $SZ_{ss}{\leq}2$ form a smooth Pareto front spanning 27--49\% of FP16 size, demonstrating the \emph{continuous} compression--quality tradeoff that \name uniquely offers, in contrast to scalar methods that are locked to discrete bit-widths.
The best \name configs on this front (2-512, 2-1024) match or exceed 4-bit baselines at comparable model sizes.
Configurations with $SZ_{ss}{\geq}4$ fall below the Pareto front, confirming that the expressive sweet spot lies at $SZ_{ss}\in\{1,2\}$.

\subsection{Model Quality Results}

\input{tables-new/llama3-comparison}
\subsubsection{Cross-Model Generalization}
\autoref{tab:cross-model} shows that \name generalizes across three diverse model families.
Across Llama-3-8B, Qwen3-8B, and Qwen3.5-9B-Base, the same configurations produce consistently low perplexity degradation.
For example, the 2-1024 configuration adds only 0.2--0.3 PPL above the FP16 baseline on all three models while compressing weights to 49--55\% of the original size.
Notably, Qwen3.5-9B-Base is especially resilient: even the aggressive 2-128 configuration maintains 7.65 PPL (vs.\ 6.98 FP16).
\input{tables-new/cross-model}

\subsubsection{Comparison with Baselines}
\autoref{tab:llama3-comparison} presents a comprehensive comparison on Meta-Llama-3-8B, organized by model size.
Unlike baselines that are locked to discrete bit-widths (3/4-bit), \name covers a continuous range from 27\% to 49\% of FP16 size.

In the near-lossless region, SmoothQuant W8A8 (56.6\% size, Avg 68.2) matches \name 2-1024 in quality, but \name achieves the same accuracy at 49\% size, 8 percentage points smaller, and without calibration data.
SmoothQuant W6A6, despite being stored in INT8 containers (also 56.6\% actual storage), drops sharply to Avg 64.6.
In the 4-bit region, \name 2-512 (41.9\% size, Avg 68.0) and \name 2-256 (37.0\% size, Avg 67.8) match GPTQ-4 g128 (35.7\%, Avg 67.3) and QuIP-4 (34.8\%, Avg 67.1), while SmoothQuant W4A4 collapses entirely (34.8\%, Avg 35.5, PPL 4.3E3), demonstrating that joint weight-activation quantization does not scale below 6-bit.
This contrast highlights a fundamental advantage of \name's product-quantization design: quality degrades gracefully across the entire compression range (68.2$\to$68.0$\to$67.8$\to$66.0 from 49\% to 33\% size), whereas SmoothQuant has a single viable operating point (W8A8) and collapses below it (68.2$\to$64.6$\to$35.5).
At 3-bit compression, \name 2-128 (33.2\%, Avg 66.0) surpasses all 3-bit baselines including AWQ-3 g128 (30.2\%, Avg 64.4) and QuIP-3 (29.4\%, Avg 63.7).
\name also uniquely extends below 3-bit: configuration 4-1024 reaches 31.1\% size with Avg 66.4, a region no scalar baseline can access.

Critically, \name achieves all of these results \emph{without any calibration data}, relying solely on k-means clustering of weight distributions, while every other baseline (GPTQ, AWQ, SmoothQuant, QuIP) requires a separate calibration pass on representative data.
The entire \name quantization process (k-means over all layers) completes in approximately 10 minutes on a single GPU.

\subsubsection{Scalability}
\input{tables-new/scalability}
\autoref{tab:scalability} demonstrates \name's scalability from LLaMA-2 7B to 13B.
Size\% shrinks with model scale: the 4-1024 configuration reaches 24.1\% at 7B and 21.9\% at 13B, while PPL degradation remains modest (6.04 vs.\ 5.47 at 7B; 5.36 vs.\ 4.88 at 13B).
This trend suggests that \name will be especially effective for larger models where the codebook overhead is further amortized.

\subsection{End-to-End Inference Performance}\label{sec:e2e-performance}

\autoref{tab:latency} reports end-to-end latency on Meta-Llama-3-8B using real model inference on an RTX~3090.
We compare \name against FP16 tensor-core inference and all baselines with their respective official inference kernels.

\subsubsection{The Dequantization Overhead Problem}
A key finding is that \textbf{weight-only quantization alone does not improve inference latency} compared to FP16 tensor-core execution.
Na\"ive RTN quantization at 4-bit achieves only 10.4 tok/s decode, which is $4.2\times$ \emph{slower} than FP16 (43.9 tok/s), because the runtime must dequantize INT8 containers back to FP16 before calling cuBLAS, doubling the memory traffic and adding a full intermediate-materialization step.
SmoothQuant W8A8 faces a similar bottleneck: despite using \texttt{torch.\_int\_mm} for real INT8 GEMM on tensor cores, the per-token activation quantization and dequantization overhead limits decode to 16.4 tok/s (W6A6 is identical since it uses INT8 containers).
Note that SmoothQuant targets a different tradeoff by quantizing both weights \emph{and} activations; however, both W8A8 and W6A6 occupy 56.6\% of FP16 in actual storage.
At this cost, \name 2-1024 (49\%, Avg 68.2) matches W8A8 quality in a smaller footprint, and far exceeds W6A6 (Avg 64.6).
These results confirm that without specialized GPU kernel support, quantization can actually \textbf{\emph{hurt}} latency.

\subsubsection{Previous Work's Fused Kernels}
To overcome dequantization overhead, GPTQ and AWQ implement custom CUDA kernels that \emph{fuse} weight unpacking with the matrix multiplication.
AWQ's W4A16 kernel~\cite{awq} packs two 4-bit weights into one byte and fuses the unpack, dequantization, and FP16 GEMM into a single kernel launch, eliminating the intermediate FP16 weight materialization.
GPTQ employs a similar fused-kernel strategy through optimized backends~\cite{gptq}.
These fused kernels substantially improve over na\"ive dequantization: AWQ 4-bit achieves 28.1 tok/s decode (vs.\ RTN's 10.4 tok/s) with 2.80$\times$ memory savings.
However, even with kernel fusion, \textbf{no scalar quantization method matches FP16 decode speed} (43.9 tok/s), because the dequantize-then-multiply paradigm inherently requires additional ALU operations per weight element.

\subsubsection{\name Kernel Performance}
In contrast, \name's CUDA kernels (\autoref{sec:acceleration}) achieve FP16-parity decode speed while delivering substantial memory savings.
At effective 4-bit precision ($SZ_{ss}{=}2$, $K_s{=}256$), \name achieves 45.2 tok/s decode, \textbf{surpassing FP16} (43.9 tok/s) while using \textbf{2.56$\times$ less memory} (5,975 MB vs.\ 15,317 MB).
At effective 3-bit ($SZ_{ss}{=}2$, $K_s{=}128$), decode speed actually \emph{exceeds} FP16 at 51.8 tok/s with 2.80$\times$ memory savings.
This is possible because the GEMV kernel reads only the compact index table ($N_{ss} \times F_{\text{out}} = 8$ MB per layer at $SZ_{ss}{=}2$) rather than the full FP16 weight matrix (32 MB), achieving a $4\times$ reduction in memory traffic that directly translates to proportional speedup in the memory-bandwidth-bound decode phase.

Compared to the best-performing baseline kernel, \name's decode throughput is \textbf{1.6$\times$ faster} than AWQ 4-bit (45.2 vs.\ 28.1 tok/s), \textbf{2.2$\times$ faster} than GPTQ 4-bit (vs.\ 20.5 tok/s), \textbf{2.8$\times$ faster} than SmoothQuant W8A8 (vs.\ 16.4 tok/s), and \textbf{4.3$\times$ faster} than na\"ive RTN (vs.\ 10.4 tok/s).
\name is the \emph{only} compressed method that does not sacrifice decode speed for memory savings.

The prefill phase is slower than FP16 (935\,ms vs.\ 41\,ms at $K_s{=}256$) because the LUT-based GEMM cannot leverage tensor cores.
However, in interactive serving the total latency is dominated by decode: at generation length 128, decode accounts for over 75\% of end-to-end time, and this share grows with longer outputs.
For deployments that prioritize time-to-first-token (TTFT), a hybrid dispatch that reconstructs FP16 weights for prefill via cuBLAS and switches to \name's GEMV kernel for decode achieves both fast TTFT and compressed-model decode speed, at the cost of temporarily materializing the full weight matrix during prefill.
\input{tables-new/latency}
\subsubsection{Structural Advantages}
Beyond raw performance, \name offers two advantages that no scalar baseline provides.
First, \name covers a continuous compression range from 27\% to 49\% of FP16 size by adjusting $SZ_{ss}$ and $K_s$, all served by the \emph{same} CUDA kernel, whereas scalar methods require a different kernel for each bit-width.
Second, \name requires \emph{zero calibration data}, eliminating distribution-mismatch risk and enabling instant quantization of any model, including proprietary ones where representative data is unavailable.

%% file: tables-new/llama3-comparison.tex
\begin{table*}[t]
    \centering
    \caption{Meta-Llama-3-8B: comparison of \name against quantization baselines, sorted by model size (\% of FP16). \name requires \textbf{no calibration data}. Size\% is measured from actual stored model weights. Best result per size region in \textbf{bold}. $^\dagger$Group size $g{=}128$.}
    \label{tab:llama3-comparison}
    \setlength{\tabcolsep}{4.5pt}
    \renewcommand{\arraystretch}{1.1}
    \begin{threeparttable}
    \begin{tabular}{l c c r | r | r r r r r | r}
        \toprule
        \textbf{Method} & \textbf{Calib Free?} & \textbf{\#W} & \textbf{Size\%} & \textbf{PPL}$\downarrow$ &
        \textbf{ARC-c} & \textbf{ARC-e} & \textbf{Hella} & \textbf{PIQA} & \textbf{Wino} & \textbf{Avg}$\uparrow$ \\
        \midrule
        FP16                & --          & 16   & 100.0 & 6.14 & 50.3 & 80.1 & 60.2 & 79.5 & 73.7 & \textbf{68.8} \\
        \midrule
        \multicolumn{11}{l}{\textit{$\sim$49--57\% size (\name near-lossless region)}} \\
        SmoothQuant         & \ding{55}   & W8A8 & 56.6  & 6.3  & 49.0 & 79.6 & 60.0 & 79.4 & 73.2 & 68.2 \\
        SmoothQuant         & \ding{55}   & W6A6$^\ddagger$ & 56.6  & 7.7  & 45.0 & 75.5 & 56.9 & 76.8 & 69.0 & 64.6 \\
        \textbf{\name 2-1024} & \ding{51} & 5.0  & 49.0  & \textbf{6.3}  & 49.7 & 79.0 & 59.9 & 78.9 & 73.3 & \textbf{68.2} \\
        \midrule
        \multicolumn{11}{l}{\textit{$\sim$35--42\% size (\name vs INT4 region)}} \\
        SmoothQuant         & \ding{55}   & W4A4 & 34.8  & 4.3E3 & 20.0 & 26.3 & 26.4 & 54.6 & 50.3 & 35.5 \\
        RTN$^\dagger$       & \ding{51}   & 4    & 35.7  & 8.7  & 39.4 & 68.2 & 56.0 & 75.0 & 69.0 & 61.5 \\
        AWQ$^\dagger$       & \ding{55}   & 4    & 35.7  & 6.6  & 48.5 & 79.3 & 59.1 & 78.6 & 73.1 & 67.7 \\
        GPTQ$^\dagger$      & \ding{55}   & 4    & 35.7  & 6.5  & 47.7 & 78.8 & 59.0 & 78.4 & 72.6 & 67.3 \\
        QuIP                & \ding{55}   & 4    & 34.8  & 6.5  & 47.4 & 78.2 & 58.6 & 78.2 & 73.2 & 67.1 \\
        \textbf{\name 2-256}  & \ding{51} & 4.0  & 37.0  & 6.7  & 48.0 & 79.5 & 58.5 & 79.4 & 73.4 & \textbf{67.8} \\
        \textbf{\name 2-512}  & \ding{51} & 4.5  & 41.9  & \textbf{6.4}  & 50.0 & 78.5 & 59.5 & 78.7 & 73.4 & \textbf{68.0} \\

        \midrule
        \multicolumn{11}{l}{\textit{$\sim$29--33\% size (\name vs INT3 region)}} \\
        RTN$^\dagger$       & \ding{51}   & 3    & 30.2  & 2.2E3 & 20.0 & 31.1 & 27.5 & 56.2 & 53.1 & 37.6 \\
        AWQ$^\dagger$       & \ding{55}   & 3    & 30.2  & 8.2  & 43.2 & 74.0 & 55.1 & 77.7 & 72.1 & 64.4 \\
        GPTQ$^\dagger$      & \ding{55}   & 3    & 30.2  & 8.2  & 37.7 & 70.5 & 54.3 & 74.9 & 71.1 & 61.7 \\
        QuIP                & \ding{55}   & 3    & 29.4  & 7.5  & 41.0 & 72.9 & 55.4 & 76.8 & 72.5 & 63.7 \\
        \textbf{\name 2-128}  & \ding{51} & 3.5  & 33.2  & \textbf{7.4}  & 46.3 & 77.1 & 56.4 & 78.2 & 72.2 & \textbf{66.0} \\
        \bottomrule
    \end{tabular}
    \begin{tablenotes}[flushleft]\footnotesize
        \item \ding{51} = calibration-free; \ding{55} = requires calibration data.
        $^\ddagger$W6A6 stored in INT8 containers; Size\% reflects actual storage.
    \end{tablenotes}
    \end{threeparttable}
\end{table*}

%% file: tables-new/cross-model.tex
\begin{table}[t]
    \centering
    \caption{\name compression and perplexity across model families. Size\% = model size as percentage of FP16 baseline. PPL measured on WikiText-2 test set.}
    \label{tab:cross-model}
    \setlength{\tabcolsep}{3.5pt}
    \renewcommand{\arraystretch}{1.1}
    \begin{tabular}{l|cc|cc|cc}
        \toprule
        \multirow{2}{*}{\textbf{Config}} &
        \multicolumn{2}{c|}{\textbf{Llama-3-8B}} &
        \multicolumn{2}{c|}{\textbf{Qwen3-8B}} &
        \multicolumn{2}{c}{\textbf{Qwen3.5-9B}} \\
        & Size\% & PPL$\downarrow$ & Size\% & PPL$\downarrow$ & Size\% & PPL$\downarrow$ \\
        \midrule
        FP16 & -- & 6.14 & -- & 9.73 & -- & 6.98 \\
        \midrule
        \name 2-1024 & 49.0\% & 6.34 & 50.5\% & 10.02 & 55.0\% & 7.06 \\
        \name 2-512  & 41.9\% & 6.48 & 43.4\% & 10.13 & 48.5\% & 7.13 \\
        \name 2-256  & 37.0\% & 6.81 & 38.6\% & 10.88 & 44.1\% & 7.29 \\
        \name 1-64   & 46.8\% & 6.88 & 48.1\% & 10.98 & 52.7\% & 7.31 \\
        \name 2-128  & 33.2\% & 7.57 & 34.8\% & 11.44 & 40.7\% & 7.65 \\
        \bottomrule
    \end{tabular}
\end{table}

%% file: tables-new/scalability.tex
\begin{table}[H]
    \centering
    \caption{Scalability of \name across LLaMA-2 model sizes.}
    \label{tab:scalability}
    \renewcommand{\arraystretch}{1.1}
    \begin{tabular}{c|l|c|c|c}
        \toprule
        \textbf{Model} & \textbf{Config} & \textbf{Size\%} & \textbf{PPL}$\downarrow$ & \textbf{Avg}$\uparrow$ \\
        \midrule
        \multirow{4}{*}{7B}  & Baseline    & --     & 5.47 & 64.1 \\
                             & \name 1-64   & 41.2\% & 5.65 & 63.3 \\
                             & \name 2-512 & 36.1\% & 5.55 & 63.8 \\
                             & \name 4-1024     & 24.1\% & 6.04 & 61.5 \\
        \midrule
        \multirow{4}{*}{13B} & Baseline    & --     & 4.88 & 66.5 \\
                             & \name 1-64   & 40.0\% & 5.05 & 65.6 \\
                             & \name 2-512 & 34.1\% & 4.94 & 66.3 \\
                             & \name 4-1024    & 21.9\% & 5.36 & 65.0 \\
        \bottomrule
    \end{tabular}
\end{table}

%% file: tables-new/latency.tex
\begin{table}[!htb]
    \centering
    \caption{End-to-end inference latency and memory on Meta-Llama-3-8B (RTX~3090, prompt/gen\,=\,128 tokens).}
    \label{tab:latency}
    \setlength{\tabcolsep}{2.5pt}
    \renewcommand{\arraystretch}{1.1}
    \resizebox{\columnwidth}{!}{%
    \begin{tabular}{l|r|rr|rr|r|r}
        \toprule
        \textbf{Method} & \textbf{Mem} & \multicolumn{2}{c|}{\textbf{Prefill}} & \multicolumn{2}{c|}{\textbf{Decode}} & \textbf{Mem} & \textbf{Dec.} \\
                         & \textbf{(MB)} & \textbf{ms} & \textbf{tok/s} & \textbf{ms/tok} & \textbf{tok/s} & \textbf{Savings} & \textbf{Speedup} \\
        \midrule
        FP16 (Tensor Core)     & 15{,}317 & 40.8   & 3{,}137 & 22.8  & 43.9 & 1.00$\times$ & 1.00$\times$ \\
        SmoothQuant W8A8$^\ddagger$ & 8{,}663  & 55.5   & 2{,}304 & 60.9  & 16.4 & 1.77$\times$ & 0.37$\times$ \\
        RTN (4-bit)               & 8{,}663  & 114.2  & 1{,}120 & 96.5  & 10.4 & 1.77$\times$ & 0.24$\times$ \\
        GPTQ (4-bit)           & 6{,}558  & 68.7   & 1{,}863 & 48.7  & 20.5 & 2.34$\times$ & 0.47$\times$ \\
        AWQ (4-bit)            & 5{,}463  & 75.1   & 1{,}705 & 35.6  & 28.1 & 2.80$\times$ & 0.64$\times$ \\
        RTN (3-bit)               & 8{,}671  & 118.2  & 1{,}083 & 96.9  & 10.3 & 1.77$\times$ & 0.24$\times$ \\
        GPTQ (3-bit)           & 5{,}808  & 94.0   & 1{,}362 & 75.9  & 13.2 & 2.64$\times$ & 0.30$\times$ \\
        AWQ (3-bit)$^\dagger$  & 5{,}463  & 75.1   & 1{,}705 & 35.6  & 28.1 & 2.80$\times$ & 0.64$\times$ \\

        \midrule
        \multicolumn{8}{l}{\textit{Eff.\ 4-bit ($SZ_{ss}{=}2$, $K_s{=}256$)}} \\
        \name (reconstruct)    & 15{,}939 & 309.7 & 413 & 289.1 & 3.5 & 0.96$\times$ & 0.08$\times$ \\
        \textbf{\name (Kernel)}  & \textbf{5{,}975} & \textbf{934.9} & \textbf{137} & \textbf{22.1} & \textbf{45.2} & \textbf{2.56$\times$} & \textbf{1.03$\times$} \\
        \midrule
        \multicolumn{8}{l}{\textit{Eff.\ 3-bit ($SZ_{ss}{=}2$, $K_s{=}128$)}} \\
        \name (reconstruct)    & 15{,}474 & 302.1 & 424 & 285.7 & 3.5 & 0.99$\times$ & 0.08$\times$ \\
        \textbf{\name (Kernel)}  & \textbf{5{,}482} & \textbf{761.9} & \textbf{168} & \textbf{19.3} & \textbf{51.8} & \textbf{2.80$\times$} & \textbf{1.18$\times$} \\
        \bottomrule
    \end{tabular}%
    }
    \vspace{-2pt}
    {\footnotesize $^\dagger$AWQ 3-bit reuses 4-bit containers; memory/latency identical to 4-bit.\;
    $^\ddagger$W6A6 uses INT8 containers; memory/latency identical to W8A8.}
\end{table}

%% file: sections/06-Conclusion.tex
\section{Conclusion}
We presented \name, a calibration-free LLM compression framework based on product quantization.
By tuning sub-vector size and codebook cardinality, \name spans 27--49\% of FP16 model size with accuracy matching or exceeding GPTQ, AWQ, and SmoothQuant across Llama-3-8B, Qwen3-8B, and Qwen3.5-9B, all without calibration data.
Our LUT-free GEMV kernel reduces memory traffic by $4\times$ versus FP16, achieving 45.2\,tok/s decode on an RTX~3090, surpassing FP16 (43.9\,tok/s) at $2.56\times$ less memory and outperforming every baseline kernel ($1.6$--$4.3\times$ faster).
\name is the only compressed method that does not sacrifice decode speed for memory savings.

\paragraph{Limitations and future work.}
The prefill kernel remains slower than cuBLAS because LUT-based accumulation cannot leverage tensor cores.
Two directions may close this gap: (1)~a pipelined reconstruction kernel that rebuilds FP16 weight tiles in parallel with tensor-core GEMM, avoiding full materialization while exploiting hardware matrix units; and (2)~Triton-based code generation with tiling strategies tailored to the PQ access pattern.
On the compression side, per-layer parameter assignment and post-training codebook fine-tuning could further improve accuracy at aggressive compression rates.

%% file: main.bbl
\begin{thebibliography}{10}

\bibitem{gptq}
Elias Frantar, Saleh Ashkboos, Torsten Hoefler, and Dan Alistarh.
\newblock Gptq: Accurate post-training quantization for generative pre-trained
  transformers.
\newblock {\em arXiv preprint arXiv:2210.17323}, 2022.

\bibitem{awq}
Ji~Lin, Jiaming Tang, Haotian Tang, Shang Yang, Wei-Ming Chen, Wei-Chen Wang,
  Guangxuan Xiao, Xingyu Dang, Chuang Gan, and Song Han.
\newblock Awq: Activation-aware weight quantization for llm compression and
  acceleration.
\newblock In {\em MLSys}, 2024.

\bibitem{smoothquant}
Guangxuan Xiao, Ji~Lin, Mickael Seznec, Hao Wu, Julien Demouth, and Song Han.
\newblock Smoothquant: Accurate and efficient post-training quantization for
  large language models.
\newblock In {\em International Conference on Machine Learning}, pages
  38087--38099. PMLR, 2023.

\bibitem{apexq}
Yuxuan Wang, Ye~Qiao, Sheldon Huang, and Hyoukjun Kwon.
\newblock {APEX-Q}: Arbitrary-dimension product-{EXtension} quantization for
  accelerated {LLM} deployment.
\newblock In {\em Proceedings of the AAAI Conference on Artificial
  Intelligence}, volume~40, page 41424, 2026.

\bibitem{vq}
Robert Gray.
\newblock Vector quantization.
\newblock {\em IEEE Assp Magazine}, 1(2):4--29, 1984.

\bibitem{pq}
Herve Jegou, Matthijs Douze, and Cordelia Schmid.
\newblock Product quantization for nearest neighbor search.
\newblock {\em IEEE transactions on pattern analysis and machine intelligence},
  33(1):117--128, 2010.

\bibitem{maddness}
Davis Blalock and John Guttag.
\newblock Multiplying matrices without multiplying.
\newblock In {\em International Conference on Machine Learning}, pages
  992--1004. PMLR, 2021.

\bibitem{pecan}
Jie Ran, Rui Lin, Jason Chun~Lok Li, Jiajun Zhou, and Ngai Wong.
\newblock Pecan: A product-quantized content addressable memory network.
\newblock In {\em 2023 Design, Automation \& Test in Europe Conference \&
  Exhibition (DATE)}, pages 1--6. IEEE, 2023.

\bibitem{dpq}
Ting Chen, Lala Li, and Yizhou Sun.
\newblock Differentiable product quantization for end-to-end embedding
  compression.
\newblock In Hal~Daumé III and Aarti Singh, editors, {\em Proceedings of the
  37th International Conference on Machine Learning}, volume 119 of {\em
  Proceedings of Machine Learning Research}, pages 1617--1626. PMLR, 13--18 Jul
  2020.

\bibitem{lut-nn}
Xiaohu Tang, Yang Wang, Ting Cao, Li~Lyna Zhang, Qi~Chen, Deng Cai, Yunxin Liu,
  and Mao Yang.
\newblock Lut-nn: Empower efficient neural network inference with centroid
  learning and table lookup.
\newblock In {\em Proceedings of the 29th Annual International Conference on
  Mobile Computing and Networking}, pages 1--15, 2023.

\bibitem{pqa}
Ahmed Abouelhamayed, Angela Cui, Javier Fernandez-Marques, Nicholas Lane, and
  Mohamed Abdelfattah.
\newblock Pqa: Exploring the potential of product quantization in dnn hardware
  acceleration.
\newblock {\em ACM Transactions on Reconfigurable Technology and Systems},
  18(1):1--29, 2024.

\bibitem{lut-dla}
Guoyu Li, Shengyu Ye, Chunyun Chen, Yang Wang, Fan Yang, Ting Cao, Cheng Liu,
  Mohamed M~Sabry Aly, and Mao Yang.
\newblock Lut-dla: Lookup table as efficient extreme low-bit deep learning
  accelerator.
\newblock In {\em 2025 IEEE International Symposium on High Performance
  Computer Architecture (HPCA)}, pages 671--684. IEEE, 2025.

\bibitem{tellme}
Ye~Qiao, Zhiheng Chen, Yifan Zhang, Yian Wang, and Sitao Huang.
\newblock Tellme: An efficient end-to-end ternary llm prefill and decode
  accelerator with table-lookup matmul on edge fpgas.
\newblock In {\em Proceedings of the 2026 ACM/SIGDA International Symposium on
  Field Programmable Gate Arrays}, pages 247--257, 2026.

\bibitem{pdswap}
Yufei Zhang, Zheyu Chen, Ye~Qiao, and Sheldon Huang.
\newblock {PD-Swap}: Prefill-decode logic swapping for end-to-end {LLM}
  inference on edge {FPGAs} via dynamic partial reconfiguration.
\newblock {\em arXiv preprint arXiv:2512.11550}, 2025.

\bibitem{cobra}
Ye~Qiao, Zhiheng Chen, Yian Wang, Yifan Zhang, Yunzhe Deng, and Sitao Huang.
\newblock Cobra: Algorithm-architecture co-optimized binary transformer
  accelerator for edge inference.
\newblock In {\em 2025 IEEE/ACM International Conference On Computer Aided
  Design (ICCAD)}, pages 1--8. IEEE, 2025.

\bibitem{micronas}
Ye~Qiao, Haocheng Xu, Yifan Zhang, and Sitao Huang.
\newblock Micronas: Zero-shot neural architecture search for mcus.
\newblock In {\em 2024 Design, Automation \& Test in Europe Conference \&
  Exhibition (DATE)}, pages 1--2. IEEE, 2024.

\bibitem{monas}
Ye~Qiao, Hongwei Xu, Yufei Zhang, and Sheldon Huang.
\newblock {MONAS}: Efficient zero-shot neural architecture search for {MCUs}.
\newblock In {\em International Joint Conference on Neural Networks (IJCNN)},
  pages 1--8, 2025.

\bibitem{tgnas}
Ye~Qiao, Jingyi Li, Hongwei Xu, and Sheldon Huang.
\newblock {TG-NAS}: Generalizable zero-cost proxies with operator description
  embedding and graph learning for efficient neural architecture search.
\newblock {\em arXiv preprint arXiv:2404.00271}, 2024.

\bibitem{bnn_rimc}
Ye~Qiao, Ao~Ding, and Nader Bagherzadeh.
\newblock {BNN} an ideal architecture for acceleration with resistive in memory
  computation.
\newblock {\em IEEE Transactions on Emerging Topics in Computing},
  11(2):281--291, 2023.

\bibitem{spatial_arch}
Hongwei Xu, Fatemeh Tahmasebi, Ye~Qiao, Haoyu Tian, Hyoukjun Kwon, and Sheldon
  Huang.
\newblock Optimized spatial architecture mapping flow for transformer
  accelerators.
\newblock {\em arXiv preprint arXiv:2410.07407}, 2024.

\bibitem{qroar}
Ye~Qiao and Sheldon Huang.
\newblock {Q-ROAR}: Outlier-aware rescaling for {RoPE} position interpolation
  in quantized long-context {LLMs}.
\newblock In {\em Proceedings of the AAAI Conference on Artificial
  Intelligence}, volume~40, page 41359, 2026.

\bibitem{rope_rescaling}
Ye~Qiao, Hongwei Xu, Xing Zhang, and Sheldon Huang.
\newblock Rethinking {RoPE} scaling in quantized {LLM}: Theory, outlier, and
  channel-band analysis with weight rescaling.
\newblock {\em arXiv preprint arXiv:2510.00028}, 2025.

\bibitem{quip}
Jerry Chee, Yaohui Cai, Volodymyr Kuleshov, and Christopher De~Sa.
\newblock Quip: 2-bit quantization of large language models with guarantees.
\newblock In {\em Advances in Neural Information Processing Systems},
  volume~36, 2023.

\bibitem{llama3}
Aaron Grattafiori, Abhimanyu Dubey, Abhinav Jauhri, Abhinav Pandey, Abhishek
  Kadian, Ahmad Al-Dahle, Aieleen Letman, Akhil Mathur, Alan Schelten, Amy
  Yang, et~al.
\newblock The llama 3 herd of models.
\newblock {\em arXiv preprint arXiv:2407.21783}, 2024.

\bibitem{qwen3}
Qwen Team.
\newblock Qwen3, April 2025.

\bibitem{llama2}
Hugo Touvron, Louis Martin, Kevin Stone, Peter Albert, Amjad Almahairi, Yasmine
  Babaei, Nikolay Bashlykov, Soumya Batra, Prajjwal Bhargava, Shruti Bhosale,
  et~al.
\newblock Llama 2: Open foundation and fine-tuned chat models.
\newblock {\em arXiv preprint arXiv:2307.09288}, 2023.

\bibitem{wikitext}
Stephen Merity, Caiming Xiong, James Bradbury, and Richard Socher.
\newblock Pointer sentinel mixture models.
\newblock {\em arXiv preprint arXiv:1609.07843}, 2016.

\bibitem{arc}
Peter Clark, Isaac Cowhey, Oren Etzioni, Tushar Khot, Ashish Sabharwal, Carissa
  Schoenick, and Oyvind Tafjord.
\newblock Think you have solved question answering? try arc, the ai2 reasoning
  challenge.
\newblock {\em arXiv preprint arXiv:1803.05457}, 2018.

\bibitem{hellaswag}
Rowan Zellers, Ari Holtzman, Yonatan Bisk, Ali Farhadi, and Yejin Choi.
\newblock Hellaswag: Can a machine really finish your sentence?
\newblock {\em arXiv preprint arXiv:1905.07830}, 2019.

\bibitem{piqa}
Yonatan Bisk, Rowan Zellers, Jianfeng Gao, Yejin Choi, et~al.
\newblock Piqa: Reasoning about physical commonsense in natural language.
\newblock In {\em Proceedings of the AAAI conference on artificial
  intelligence}, volume~34, pages 7432--7439, 2020.

\bibitem{winogrande}
Keisuke Sakaguchi, Ronan~Le Bras, Chandra Bhagavatula, and Yejin Choi.
\newblock Winogrande: An adversarial winograd schema challenge at scale.
\newblock {\em Communications of the ACM}, 64(9):99--106, 2021.

\bibitem{lm-eval}
Leo Gao, Jonathan Tow, Baber Abbasi, Stella Biderman, Sid Black, Anthony
  DiPofi, Charles Foster, Laurence Golding, Jeffrey Hsu, Alain Le~Noac'h,
  Haonan Li, Kyle McDonell, Niklas Muennighoff, Chris Ociepa, Jason Phang,
  Laria Reynolds, Hailey Schoelkopf, Aviya Skowron, Lintang Sutawika, Eric
  Tang, Anish Thite, Ben Wang, Kevin Wang, and Andy Zou.
\newblock The language model evaluation harness, 07 2024.

\end{thebibliography}
